\documentclass[conference,table]{IEEEtran}
\IEEEoverridecommandlockouts
\usepackage{bbding}
\usepackage{cite}
\usepackage{amsmath,amssymb,amsfonts}
\usepackage{bm} 
\usepackage{textcomp}
\usepackage{caption}
\usepackage{graphicx}
\usepackage{pgfplots}
\pgfplotsset{compat=1.18}
\usepackage{subcaption}
\usepackage{multirow}
\usepackage{algorithm}  
\usepackage{algpseudocode}  

\pgfplotsset{ 
cycle list={%
{draw=black,mark=star,solid},
{draw=black, mark=square,solid}}}
\usepackage{breqn}
\usepackage{hyperref} 
\usepackage{fancyhdr}
\usepackage{subcaption}
\usepackage{comment}
\usepackage{authblk}
\usepackage{fancyhdr}
\usepackage{subcaption}
\IEEEoverridecommandlockouts 

\usepackage{fancyhdr}
\fancypagestyle{firstpage}
 {
     \fancyhead[L]{\footnotesize © 2026 IEEE.  Personal use of this material is permitted.  Permission from IEEE must be obtained for all other uses, in any current or future media, including reprinting/republishing this material for advertising or promotional purposes, creating new collective works, for resale or redistribution to servers or lists, or reuse of any copyrighted component of this work in other works. This paper is accepted at the  IEEE Latin-American Test Symposium (LATS) 2026.}
     \fancyhead[R]{}
 }
\begin{document}

\title{RESQ: A Unified Framework for \underline{RE}liability- and \underline{S}ecurity Enhancement of \underline{Q}uantized Deep Neural Networks}

\author[1]{Ali Soltan Mohammadi}
\author[1]{Samira Nazari}
\author[1]{Ali Azarpeyvand}
\author[2,3]{Mahdi Taheri}
\author[4]{Milos Krstic}
\author[2]{\\Michael Hübner}
\author[2]{Christian Herglotz}
\author[3]{Tara Ghasempouri}

\affil[1]{University of Zanjan, Zanjan, Iran}
\affil[2]{Brandenburg Technical University, Cottbus, Germany}
\affil[3]{Tallinn University of Technology, Tallinn, Estonia}
\affil[4]{Leibniz Institute for High Performance Microelectronics, Frankfurt Oder, Germany}

\maketitle
\thispagestyle{firstpage}

\begin{abstract}
This work proposes a unified three-stage framework that produces a quantized DNN with balanced fault and attack robustness. The first stage improves attack resilience via fine-tuning that desensitizes feature
representations to small input perturbations. The second stage reinforces fault resilience through fault-aware fine-tuning under simulated bit-flip faults. Finally, a lightweight post-training adjustment integrates quantization to enhance efficiency and further mitigate fault sensitivity without degrading attack resilience. Experiments on ResNet18, VGG16, EfficientNet, and Swin-Tiny in CIFAR-10, CIFAR-100, and GTSRB show consistent gains of up to 10.35\% in attack resilience and 12.47\% in fault resilience, while maintaining competitive accuracy in quantized networks. The results also highlight an asymmetric interaction in which improvements in fault resilience generally increase resilience to adversarial attacks, whereas enhanced adversarial resilience does not necessarily lead to higher fault resilience.

\end{abstract}


\section{Introduction}

Deploying DNNs on edge platforms for safety-critical applications introduces new concerns related to reliability and security. Despite their high inference accuracy,
DNNs are vulnerable to two primary threats: (1) adversarial
attacks, where imperceptible input perturbations can mislead
models into incorrect predictions \cite{khamaiseh2022adversarial}, and (2) hardware-induced
faults, such as Single Event Upsets (SEUs) that can cause bitflips in memory, which can arise from cosmic radiation \cite{mittal2020survey}.
These vulnerabilities threaten prediction integrity, particularly
in safety-critical contexts like autonomous vehicles, industrial
robotics, and defense navigation systems, where adversarial
manipulation of sensor readings or bit-flip faults can lead to
hazardous actions and misclassifications.

At the same time, Quantized Deep Neural Networks (QNNs) have emerged as a compelling solution for resource-constrained edge environments \cite{rokh2023comprehensive}. By representing weights and activations with low-precision formats, quantization reduces memory and energy consumption. However, reducing precision also introduces numerical instabilities, such as larger relative changes in weight values after small perturbations, increased sensitivity to bit flips in the most significant bits, and abrupt shifts caused by quantization-level boundaries. These effects make QNNs more vulnerable to both adversarial perturbations and hardware-induced faults \cite{sze2017efficient}. Hence, the interplay between quantization, fault resilience, and attack resilience becomes critical for dependable deployment.

Substantial efforts focus on improving the reliability of DNNs under hardware-induced faults. Prior studies highlight that traditional resilience techniques fail to address the unique dataflow and error-propagation patterns of DNNs ~\cite{mittal2020survey}. Reliability-aware retraining approaches and sensitivity-guided fine-tuning methods show improved tolerance to weight and activation bit-flips. Other studies explore selective protection or fault-aware quantization~\cite{nazari2024fortune}, as well as hybrid mitigation strategies that adjust model parameters to reduce vulnerability under soft errors~\cite{mittal2020survey}. These representative efforts collectively underscore the growing need for reliability-aware design and deployment in safety-critical environments.

Parallel to fault resilience, extensive efforts have been devoted to mitigating adversarial vulnerabilities\cite{rahman2025evaluating} in DNNs.  In this context, adversarial threats primarily refer to test-time evasion attacks, where small, imperceptible perturbations are added to the input data to mislead the model. 
A large body of work~\cite{eleftheriadis2024adversarial} analyze attack strategies such as the Fast Gradient Sign Method (FGSM)\cite{villegas2024evaluating}, and Momentum Iterative Method (MIM)\cite{li2024survey}, as well as corresponding defense mechanisms including adversarial training, gradient masking, and input denoising. Recent advancements in Vision Transformers (ViTs) have improved visual recognition in traffic sign detection for autonomous driving, yet their vulnerability in adversarial environments necessitates a comprehensive review of their weaknesses and robustness enhancement methods, complemented by a detailed tabular comparison\cite{fawole2025recent}.

Despite these advancements, most existing studies treat faults and attacks as isolated challenges, overlooking their potential interactions and combined effects. This separation leaves open important questions about how techniques designed to address one vulnerability might affect the other. Our findings indicate that adversarial fine-tuning can reduce fault tolerance and that fault-aware training does not always preserve adversarial robustness. This observation motivates the need for a unified framework that addresses both forms of resilience while maintaining computational efficiency.

To address this challenge, this paper proposes a unified three-stage framework that sequentially improves both fault and attack resilience, culminating in a quantized, dual-resilient DNN suitable for edge deployment.

The key contributions of this paper are summarized as follows:

\begin{itemize}
    \item \textbf{Dual-Resilience Analysis:} A systematic investigation of the interdependence between fault resilience and attack resilience in DNNs, revealing their asymmetric trade-off characteristics.
    
    \item \textbf{Unified Sequential Framework:}   A unified resilience-enhancement pipeline that simultaneously strengthens resilience against adversarial perturbations applied to the input and improves tolerance to bit-flip faults in memory, delivering balanced protection across both threat domains.
    
    \item \textbf{Quantized Resilient DNN Output:} The framework produces a quantized model that maintains high accuracy while exhibiting enhanced resilience to both hardware faults and adversarial attacks.
    
    \item \textbf{Cross-Architecture Evaluation:} Validation on ResNet18, VGG16, EfficientNet, and Swin-Tiny across CIFAR-10 and GTSRB datasets demonstrates the generality of the proposed method across both CNNs and ViTs.
\end{itemize}

The remainder of this paper is organized as follows. Section~\ref{sec:methodology} presents the proposed three-stage methodology for enhancing fault and attack resilience. Section~\ref{sec:results} discusses the experimental setup, evaluation metrics, and key results across multiple DNN architectures and datasets. Finally, Section~\ref{sec:conclusion} concludes the paper.

\section{Proposed Methodology}
\label{sec:methodology}

This section introduces RESQ, a unified framework designed to achieve fault- and attack-resilient DNNs. RESQ integrates complementary resilience-enhancement stages that strengthen the model against adversarial perturbations and memory bit-flip faults. The framework combines adversarial attack resilience, improved using Bit Plane Feature Consistency (BPFC)~\cite{addepalli2020towards}, with fault resilience achieved through fault-aware fine-tuning and post-training protection based on FORTUNE~\cite{nazari2024fortune}. Unlike prior works that address these vulnerabilities independently, RESQ sequentially couples resilience-enhancing steps to achieve balanced protection without degrading model accuracy. The final output is a quantized DNN that remains resilient against both adversarial input manipulations and hardware-induced faults.

\begin{figure}[ht]
    \centering
    \includegraphics[width=0.5\textwidth]{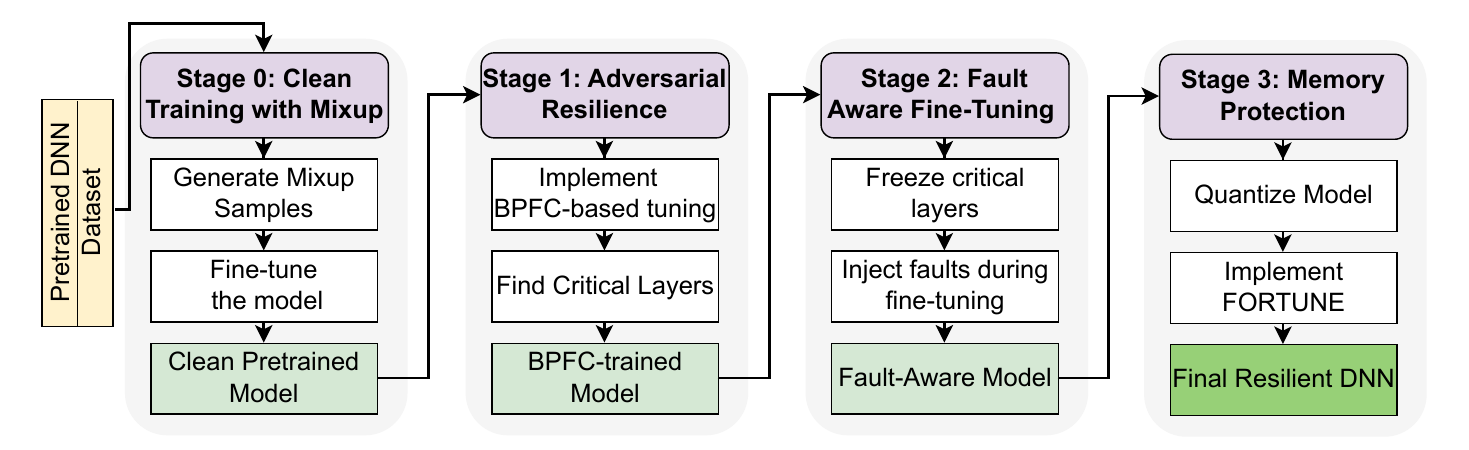}
    \caption{Overview of the RESQ framework. The pipeline sequentially enhances DNN resilience and produces a final model robust to both input perturbations and hardware faults.}
    \label{method}
\end{figure}

\subsection{Overview}

RESQ consists of one preparation stage followed by three resilience-oriented stages (Figure~\ref{alg:unified}). Stage~0 trains a strong baseline model, and Stages~1--3 progressively enhance adversarial resilience, fault tolerance, and memory-level reliability.

\begin{enumerate}
    \item \textbf{Stage 0 — Clean Training with Mixup:}
    A clean pretrained model is obtained using mixup augmentation to improve generalization and stabilize later resilience training.

    \item \textbf{Stage 1 — Adversarial Resilience:}
    The network learns to suppress sensitivity to LSB perturbations using BPFC, reducing vulnerability to adversarial attacks that operate on visually imperceptible bit changes.

    \item \textbf{Stage 2 — Fault-Aware Fine-Tuning:}
    Layers most critical for adversarial robustness are frozen, while the remaining layers are fine-tuned under simulated bit-flip faults to improve tolerance to hardware errors.

    \item \textbf{Stage 3 — Memory Protection:}
    The model is quantized using an adaptive bit-width search, and MSBs are protected via triple modular redundancy for reliable deployment.
\end{enumerate}

Throughout this section, $\Theta$ denotes trainable parameters and $f_\theta$ the corresponding model.

\begin{algorithm}[htbp]
\caption{Integrate Pipeline for Attack- and Fault-Resilient DNN Deployment}
\label{alg:unified}
\begin{algorithmic}[1]  
\Require Dataset $\mathcal{D}$, learning rate $\eta$, mixup parameter $\alpha$, LSB removal factor $k$, fault injection rate $BER$, quantization range $[m,n]$, accuracy threshold $a$, reliability threshold $r$
\Ensure Final resilient DNN $f_{\theta^*}$

\Statex \textbf{// Stage 0: Clean Training}
\For{each minibatch $B$} 
    \State Generate mixup samples $(\tilde{x},\tilde{y})$ 
    \State Update $\Theta \gets \Theta - \eta \nabla_\Theta CE(f_\theta(\tilde{x}), \tilde{y})$
\EndFor
\State Obtain $\Theta_{\text{clean}}$

\Statex \textbf{// Stage 1: Adversarial Resilience}
\For{each minibatch $B$}
    \For{each $(x_i,y_i)$}
        \State $x^{pre}_i = x_i + U(-2^{k-2}, 2^{k-2})$
        \State $x^q_i = x^{pre}_i - (x^{pre}_i \bmod 2^k)$
        \State Compute BPFC loss
    \EndFor
    \State Update $\theta$
\EndFor
\State Obtain $\Theta_{\text{BPFC}}$

\Statex \textbf{// Stage 2: Fault-Aware Fine-Tuning}
\State Identify critical layers $L_c$ via gradient EMA analysis
\State Freeze parameters of $L_c$
\For{each minibatch $B$}
    \State Inject bit-flip faults in unfrozen layers at rate $BER$
    \State Update only non-critical layers
\EndFor
\State Obtain $\Theta_{\text{FA}}$

\Statex \textbf{// Stage 3: Memory Protection}
\State Initialize bit width $b = (m+n)/2$
\While{not converged}
    \State Quantize weights (affine mapping)
    \State Evaluate accuracy and reliability
    \If{accuracy $\ge a$ and reliability $\ge r$}
        \State Apply MSB triple modular redundancy
        \State \textbf{break}
    \Else
        \State Adjust $b$ via binary search
    \EndIf
\EndWhile

\State \Return $f_{\theta^*}$
\end{algorithmic}
\end{algorithm}

\subsection{Stage 0: Clean Training with Mixup}

Before adversarial and fault-resilience training stages are applied, an initial clean training phase is conducted to establish a strong baseline model (Algorithm~\ref{alg:unified}, Lines~1--5). During this stage, the network is trained on the clean dataset using data augmentation strategies designed to improve generalization. In particular, the mixup technique is employed, where pairs of training samples $(x_i, y_i)$ and $(x_j, y_j)$ are linearly interpolated to form augmented examples:
\[
\tilde{x} = \lambda x_i + (1 - \lambda)x_j, \qquad
\tilde{y} = \lambda y_i + (1 - \lambda)y_j,
\]
where $\lambda \sim \text{Beta}(\alpha, \alpha)$. The parameter $\alpha$ denotes the mixup hyperparameter that controls the strength of interpolation, while the Beta distribution determines how strongly pairs of samples are blended. Model parameters $\Theta$ are updated using the cross-entropy loss with the predefined learning rate $\eta$.

This augmentation strategy encourages the model to learn smoother decision boundaries and reduces overfitting by exposing it to a broader space of input variations. In this stage, validation is performed to monitor convergence and guide hyperparameter choices such as the learning rate, regularization strength, and batch composition. Standard validation metrics (e.g., accuracy or validation loss) ensure that the resulting pretrained model exhibits strong generalization before resilience-oriented stages are applied. The output of Stage~0, denoted $\Theta_{\text{clean}}$, serves as the initialization for the subsequent BPFC-based adversarial resilience stage.

\subsection{Stage 1: Adversarial Resilience}

Adversarial perturbations typically exploit imperceptible variations in the Least Significant Bits (LSBs) of input images. Modifying only LSBs keeps the perturbations visually undetectable to humans, whereas altering higher-significance bits would introduce visible artifacts. To reduce the model's sensitivity to such small but adversarially crafted changes, the BPFC approach~\cite{addepalli2020towards} encourages the network to rely on more stable, high-magnitude information present in higher bit planes.

In this stage, the model is trained to maintain feature consistency between each original input $x_i$ and a quantized version $x^{q}_i$ obtained after removing $k$ LSBs (Algorithm~\ref{alg:unified}, Lines~6--14). Before bit-plane removal, a small amount of uniform noise is added,
\[
x^{pre}_i = x_i + U(-2^{k-2},\, 2^{k-2}),
\]
to prevent degenerate solutions and encourage resilience across local neighborhoods of the input space. The quantized variant is then computed as
\[
x^{q}_i = x^{pre}_i - (x^{pre}_i \bmod 2^k),
\]
which eliminates the lowest $k$ bits of each pixel.

To enforce invariance between $x_i$ and $x^{q}_i$, training minimizes the BPFC loss:
\[
\mathcal{L}_{BPFC}
= CE(f_\theta(x_i), y_i)
+ \lambda \bigl\| g(x_i) - g(x^{q}_i) \bigr\|_2^2,
\]
where $\lambda$ controls the strength of the feature-consistency regularization and $g(\cdot)$ denotes the pre-softmax activations and $CE$ is the cross entropy loss. This regularizer reduces the model's local Lipschitz constant, leading to smoother decision boundaries and substantially improved resistance to adversarial perturbations. The updated parameters are obtained using the learning rate $\eta$, and the resulting model $\Theta_{\text{BPFC}}$ serves as the initialization for the subsequent fault-resilience stage.

\subsection{Stage 2: Fault-Aware Fine-Tuning}

Adversarial and fault-resilience objectives can interfere with each other when trained simultaneously. In particular, injecting weight perturbations during fault-aware training may degrade the feature invariances learned during BPFC. To prevent this conflict, Stage~2 employs a selective layer-freezing strategy that protects the most influential layers—those essential for adversarial robustness—while allowing controlled adaptation in less sensitive parts of the network.

\subsubsection{Identification of Critical Layers}

Critical layers are identified using a sensitivity analysis performed on the BPFC-trained model $\theta_{\text{BPFC}}$ (Algorithm~\ref{alg:unified}, Line~15). During the final iterations of Stage~1, the gradient norms of each layer are monitored to estimate their relative importance. Layers whose parameters consistently exhibit large gradients contribute more strongly to the loss landscape and are therefore considered essential for maintaining adversarial resilience.

To obtain stable estimates, an Exponential Moving Average (EMA) of gradient norms is maintained for each layer. For a layer $l$, the EMA at iteration $t$ is computed as:
\[
\text{EMA}_l^{(t)}
= \beta \, \|\nabla_{\theta_l} \mathcal{L}^{(t)}\|_2
  + (1 - \beta)\, \text{EMA}_l^{(t-1)},
\]
where $\beta \in (0,1)$ is a smoothing factor (distinct from the mixup parameter used in Stage~0), and $\|\nabla_{\theta_l} \mathcal{L}^{(t)}\|_2$ is the $\ell_2$-norm of the gradient of the loss with respect to layer $l$ at iteration $t$. A threshold is applied to these importance scores to classify layers as critical. Layers whose normalized EMA scores exceed this threshold form the set of protected layers $L_c$.

\subsubsection{Fault-Aware Fine-Tuning}

After identifying the critical layers, their parameters are \emph{frozen} during Stage~2 (Algorithm~\ref{alg:unified}, Line~16). Freezing means that the weights of these layers are excluded from gradient updates: they remain fixed throughout fault-aware fine-tuning. This preserves the BPFC-induced invariances that are crucial for adversarial robustness.

For all unfrozen (non-critical) layers, random bit-flip faults are injected during forward passes at a predefined Bit Error Rate (BER) (Algorithm~\ref{alg:unified}, Lines~17--21). The BER determines the expected fraction of bits in the weight representation that are flipped. For a weight $w_i$, the injected fault is modeled as
\[
\hat{w}_i = w_i \oplus \mathcal{B}(BER),
\]
where $\mathcal{B}(BER)$ is a Bernoulli-distributed random bit mask with parameter $BER$, and $\oplus$ denotes a bitwise XOR operation.

Training optimizes a composite loss that encourages both correct classification and robustness to weight faults:
\[
\mathcal{L}_{\text{fault}}
= \mathcal{L}_{\text{CE}}(f(W, x), y)
+ \lambda\,\mathbb{E}_{\hat{W}}
    \left[
        \| f(W,x) - f(\hat{W},x) \|_2^2
    \right],
\]
where $f(W,x)$ is the network output under nominal weights (i.e., the clean, fault-free parameters), and $f(\hat{W},x)$ is the output under fault-injected weights. The expectation operator $\mathbb{E}_{\hat{W}}[\cdot]$ denotes averaging over multiple random fault realizations. The second term penalizes deviations caused by bit flips, encouraging the non-critical layers to absorb fault resilience without disturbing the preserved adversarially robust layers.

\subsection{Stage 3: Memory Protection}

The final stage applies a modified version of FORTUNE~\cite{nazari2024fortune} to quantize and harden the model weights for reliable deployment. The algorithm performs a search over the quantization bit width $b$ (Algorithm~\ref{alg:unified}, Lines~22--33), which is initialized using the midpoint of the user-defined quantization range $[m,n]$. For each candidate bit width, the weights are quantized using an affine linear mapping:
\[
q_i = \text{round}\!\left(\frac{x_i - x_{\min}}{s}\right), 
\qquad 
s = \frac{x_{\max} - x_{\min}}{2^b - 1},
\]
where $s$ is the quantization scale. This mapping produces an unsigned integer representation in which all weights are non-negative, effectively eliminating the sign bit and improving resilience to bit-flip faults.

To enhance robustness further, the most significant bits (MSBs) of the quantized weights are protected using a triple modular redundancy (TMR) mechanism. For each weight element $i$, the protected MSB is replicated three times, forming copies $\{b_{i1}, b_{i2}, b_{i3}\}$. During inference, the MSB is reconstructed via majority voting:
\[
\tilde{b}_i = \text{mode}(b_{i1}, b_{i2}, b_{i3}) =
\begin{cases}
1, & \text{if } b_{i1} + b_{i2} + b_{i3} \geq 2, \\[3pt]
0, & \text{otherwise}.
\end{cases}
\]
Here $\text{mode}(\cdot)$ denotes majority voting, and the number of protected bits $n_{\text{MSB}}$ is a configurable parameter that allows trading memory overhead for reliability.

The algorithm evaluates, for each bit width, both model accuracy and fault tolerance under simulated bit-flip errors. Two thresholds are used for this search: (i) an accuracy requirement $a$ and (ii) a reliability requirement $r$ (Algorithm~\ref{alg:unified}, Lines~26--28). If both criteria are satisfied, TMR protection is applied and the search terminates; otherwise, the bit width $b$ is adjusted via binary search. The term “not converged” refers to the iterative search over $b$ until a configuration meeting both thresholds is found.

Through this combination of quantization, sign-bit elimination, and MSB redundancy, the final model $f_{\theta^*}$ achieves strong tolerance against memory faults while preserving the adversarial robustness and clean accuracy obtained in earlier stages.

\section{Experimental Results and Discussion}
\label{sec:results}

\subsection{Experimental Setup}

Four representative neural network architectures are employed to validate the generality of the proposed approach: ResNet-18, VGG-11, EfficientNet-B0, and Swin-Tiny. ResNet-18 incorporates residual connections to mitigate vanishing gradient issues; VGG-11 adopts a uniform convolutional structure with $3\times3$ filters; EfficientNet-B0 leverages compound scaling for optimized depth-width-resolution balance; and Swin-Tiny represents a hierarchical Vision Transformer model with shifted window self-attention. These architectures collectively enable comprehensive evaluation across both CNN- and ViT-based designs. All experiments are conducted on a computing system equipped with an NVIDIA GeForce GTX 1050 Ti GPU.

Three benchmark datasets are utilized for evaluation: CIFAR-10, CIFAR-100, and the German Traffic Sign Recognition Benchmark (GTSRB). The CIFAR-10 dataset comprises 60,000 color images of size $32\times32$ distributed across 10 classes, whereas CIFAR-100 extends this benchmark to 100 fine-grained classes. The GTSRB dataset, containing over 50,000 images from 43 traffic sign categories, is selected to assess the framework’s applicability to safety-critical perception tasks, such as those encountered in autonomous driving.

Table~\ref{tab:baseline_stage_accuracy} summarizes the baseline accuracies of the original pretrained models and the improvements obtained at each stage of the proposed unified framework—namely, (i) adversarial fine-tuning (BPFC stage), (ii) fault-aware fine-tuning (FA stage), and (iii) quantization with FORTUNE-based reliability enhancement (Q-FA stage). The results demonstrate that the proposed pipeline maintains competitive clean accuracy while progressively enhancing both attack and fault resilience.

\begin{table*}[htbp]
\centering
\caption{Baseline and Stage-Wise Accuracy (\%) of RESQ across Architectures and Datasets}
\label{tab:baseline_stage_accuracy}
\begin{tabular}{lccccc}
\hline
\textbf{Model} & \textbf{Dataset} & \textbf{Baseline} & \textbf{BPFC Stage} & \textbf{FA Stage}& \textbf{Q-FA Stage}\\
\hline
ResNet-18 & CIFAR-10 &92.82 &92.13 &91.34 &91.69 \\
VGG-11 & CIFAR-10 &90.44 &90.05 &89.26 &89.83 \\
EfficientNet-B0 & GTSRB &99.89 &99.73 &99.05 &96.40\\
Swin-Tiny & CIFAR-100 &85.30 &81.39 &78.20 &73.12\\
\hline
\end{tabular}
\end{table*}

\subsection{Layer-Wise Criticality Analysis}

\begin{figure}[htbp]
\subfloat{\includegraphics[width=0.5\textwidth]{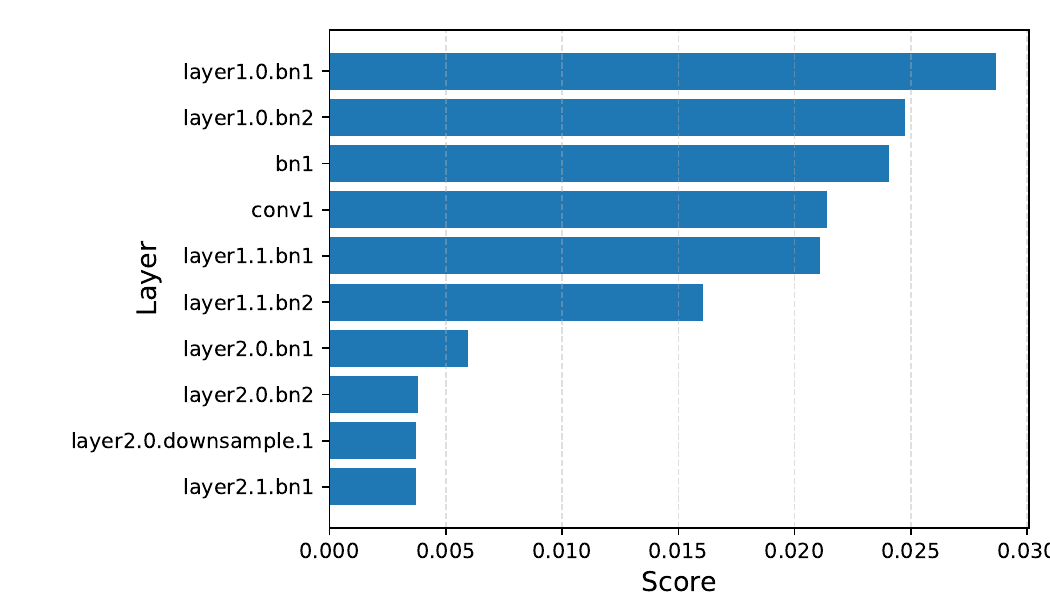}} \\
\caption{Top ten critical layers for ResNet-18 via EMA-based gradient norms.}
\label{fig:critical_layers_results}
\end{figure}

Figure~\ref{fig:critical_layers_results} presents the results of the critical layer identification procedure. Due to space limitations, only the ResNet model is shown as a representative example. The figure illustrates that the most critical layers—identified via EMA of gradient norms—are predominantly located in the early residual blocks. These initial blocks form the core feature-extraction pathway and play a major role in preserving the adversarial invariances learned during BPFC. Because of their elevated criticality scores, these early-stage layers are frozen during the fault-aware fine-tuning phase. This ensures that their adversarially resilient representations remain intact, while fault resilience is learned primarily in the non-critical layers. Consequently, the model adapts to bit-flip faults without disrupting the sensitive, robustness-preserving components of the network.

\begin{figure}[ht]
\centering

\begin{subfigure}{0.9\linewidth}
    \centering
    \includegraphics[width=\linewidth]{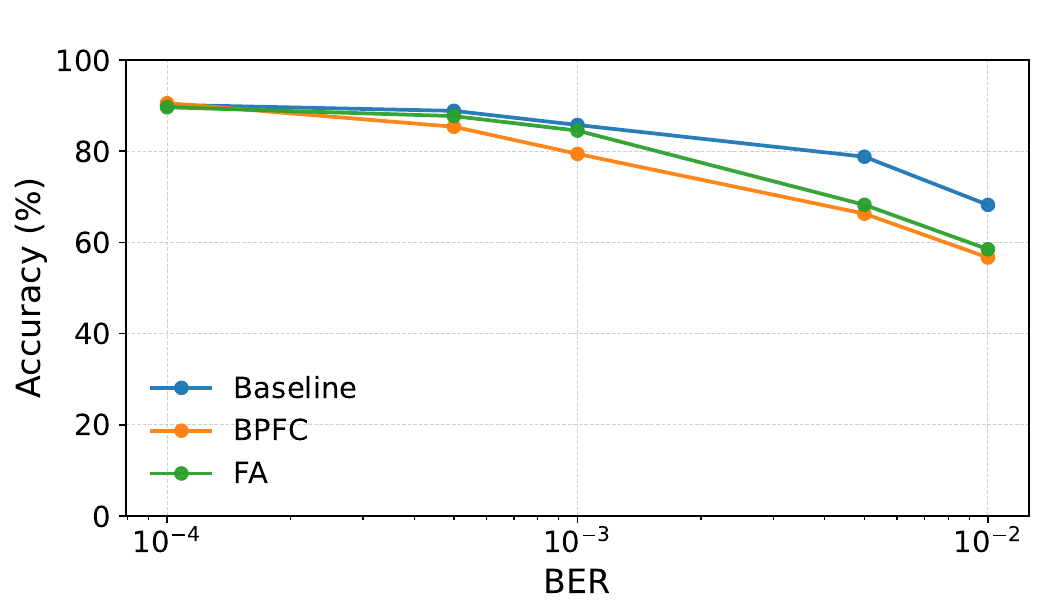}
    \caption{Fault Resilience}
    \label{fig:bpfc_sub1}
\end{subfigure}

\begin{subfigure}{0.9\linewidth}
    \centering
    \includegraphics[width=\linewidth]{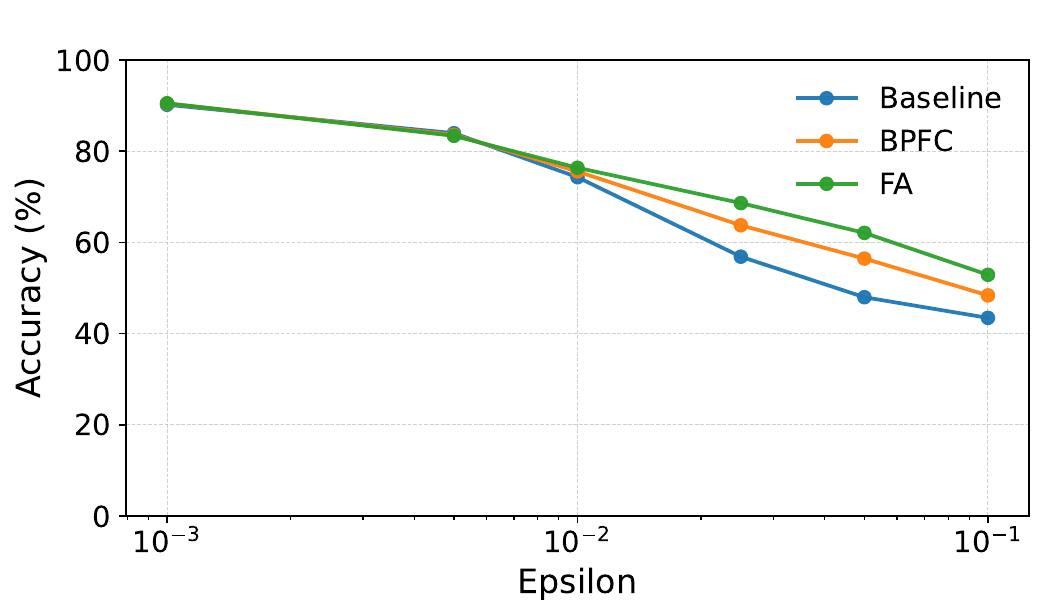}
    \caption{Attack Resilience}
    \label{fig:bpfc_sub2}
\end{subfigure}

\caption{Impact of training stages on resilience of ResNet-18}
\vspace{-0.7cm}
\label{fig:bpfc_fa_results}
\end{figure}

Following the layer-wise criticality analysis, Figure~\ref{fig:bpfc_fa_results} further illustrates how each training stage affects model resilience, shown here for ResNet-18 as a representative example. The figure contains two subplots comparing the baseline, BPFC-trained, and fault-aware trained (FA) models.

Figure~\ref{fig:bpfc_fa_results}(a) reports accuracy under injected bit-flip faults. The BPFC model exhibits the lowest fault tolerance, reflecting the trade-off where enhancing adversarial resilience increases sensitivity to weight perturbations. Fault-aware fine-tuning mitigates this effect: although the FA model does not fully recover the baseline’s fault resilience, it significantly improves over BPFC.

Figure~\ref{fig:bpfc_fa_results}(b) evaluates resilience against FGSM attacks across multiple $\epsilon$ values. Here, BPFC already provides substantial resilience gains compared to the baseline, and the FA model further strengthens this adversarial resilience. This outcome highlights the Asymmetric Resilience Observation: while BPFC improves attack resilience at the cost of fault tolerance, fault-aware training simultaneously enhances both dimensions by protecting critical layers and adapting the remaining ones to faults. Overall, the combined results demonstrate that RESQ progressively balances and strengthens both fault and attack resilience across stages.

\subsection{Evaluation on Adversarial Resilience}

The final quantized fault- and attack-resilient models (Q-FA models) are compared against corresponding quantized original baselines under multiple adversarial settings, including FGSM, IFGSM, PGD, BIM, and MIM attacks. 

Due to differences in model architectures and their intrinsic resilience to adversarial perturbations, different perturbation magnitudes $\epsilon$ are used for evaluation: $\epsilon = 0.1$ for VGG11 and ResNet18, $\epsilon = 0.001$ for EfficientNet, and $\epsilon = 0.0005$ for Swin-Tiny. Larger $\epsilon$ values are suitable for less resilient models like VGG11 and ResNet18, while more robust architectures such as EfficientNet and Swin-Tiny require smaller perturbations to effectively evaluate adversarial reilience. Table~\ref{tab:adv_all} summarizes the adversarial resilience of all models under the respective perturbation settings.

\begin{table*}[htbp]
\centering
\caption{Adversarial Resilience of Models with Different Perturbation Magnitudes}
\label{tab:adv_all}
\begin{tabular}{lcccccc|c}
\hline
\textbf{Model} & \textbf{Clean} & \textbf{FGSM} & \textbf{IFGSM} & \textbf{PGD} & \textbf{BIM} & \textbf{MIM} & \textbf{$\epsilon$}\\
\hline
VGG11 & 89.83 & 36.92 & 10.22 & 12.28 & 10.90 & 16.33 & 0.1  \\
Q-FA-VGG11 & 90.02 & 50.94 & 24.52 & 27.57 & 25.57 & 30.77 & 0.1  \\
ResNet18 & 91.69 & 41.95 & 10.52 & 12.68 & 11.06 & 17.43 & 0.1  \\
Q-FA-ResNet18 & 92.82 & 52.30 & 30.45 & 31.01 & 30.86 & 34.80 & 0.1  \\
\hline
EfficientNet & 96.40 & 58.39 & 51.47 & 46.57 & 51.65 & 54.44 & 0.001  \\
Q-FA-EfficientNet & 97.80 & 86.35 & 84.97 & 83.55 & 85.01 & 85.57 & 0.001  \\
\hline
Swin-Tiny & 72.07 & 33.80 & 30.65 & 28.03 & 30.42 & 32.02 & 0.0005  \\
Q-FA-SwinTiny & 73.74 & 35.29 & 31.05 & 27.52 & 31.07 & 32.97 & 0.0005  \\
\hline
\end{tabular}
\vspace{-0.3cm}
\end{table*}

The results show that the Q-FA models consistently improve adversarial resilience compared to the corresponding baseline models. For the less resilient architectures, VGG11 and ResNet18, which are evaluated with a larger perturbation $\epsilon = 0.1$, the improvements are substantial across all attack types, particularly FGSM, IFGSM, and MIM attacks. EfficientNet, evaluated with a smaller perturbation $\epsilon = 0.001$ due to its higher intrinsic resilience, also benefits greatly from Q-FA quantization, achieving significant accuracy gains under all attacks. Swin-Tiny, evaluated with the smallest perturbation $\epsilon = 0.0005$, shows moderate improvements, particularly for PGD and MIM attacks, although the gains are less pronounced compared to the other architectures. These results demonstrate that the proposed Q-FA approach effectively enhances adversarial resilience while accounting for differences in architecture sensitivity and inherent robustness.

\subsection{Evaluation on Fault Resilience}

To evaluate fault tolerance, random bit-flip injections are applied to model weights at varying BERs, and the resulting classification accuracy quantifies model reliability. Different BER ranges are selected based on the structural characteristics and inherent resilience of each architecture. Models such as VGG11 and EfficientNet exhibit lower natural tolerance to bit errors; therefore, smaller BER ranges are used to clearly capture their degradation patterns. In contrast, ResNet18 and Swin-Tiny are more resilient to weight perturbations, so higher BER ranges are required to meaningfully expose their resilience behavior. These architecture-specific BER settings ensure that differences in fault tolerance are clearly observable and comparable across models.

Table~\ref{tab:fault_all} summarizes the reliability scores for all models under the respective BER settings.

\begin{table}[ht]
\centering
\caption{Fault Resilience of Models under Varying BERs}
\label{tab:fault_all}
\begin{tabular}{lccc}
\hline
\textbf{Model} & \textbf{BER 1} & \textbf{BER 2} & \textbf{BER 3} \\
\hline
Q-FA-VGG11 & 91.21 & 49.97 & 13.95  \\
VGG11 & 89.83 & 49.22 & 10.01  \\
\hline
Q-FA-ResNet18 & 91.52 & 90.58 & 74.71  \\
ResNet18 & 79.05 & 18.92 & 9.99  \\
\hline
Q-FA-EfficientNet & 96.15 & 95.57 & 93.68  \\
EfficientNet & 89.14 & 87.31 & 86.17  \\
\hline
Q-FA-SwinTiny & 72.62 & 68.56 & 62.46  \\
SwinTiny & 71.90 & 65.90 & 44.50  \\
\hline
\end{tabular}
\end{table}

The results show that the Q-FA models consistently improve fault resilience across all evaluated BERs. For VGG11, which is evaluated at low BERs, the Q-FA variant maintains slightly higher accuracy than the baseline under all bit-flip rates, indicating enhanced reliability. ResNet18, evaluated at higher BERs, shows substantial improvements with Q-FA, especially at the highest BER of 0.01, where the accuracy increases from 9.99\% to 74.71\%. EfficientNet exhibits strong fault tolerance even at higher BERs, and the Q-FA model further enhances this resilience. Swin-Tiny, which is moderately resilient, shows notable improvements under the highest BERs, particularly improving reliability from 44.50\% to 62.46\% at BER = 0.005. These results confirm that the proposed fault- and attack-resilient quantization strategy effectively strengthens fault tolerance while accounting for architecture-specific sensitivity to bit errors. Following fault-aware fine-tuning, fault resilience is restored compared to pre-BPFC levels, and the subsequent FORTUNE quantization stage further enhances reliability by promoting redundancy in critical bits.

\section{Conclusion}
\label{sec:conclusion}

This work proposes a unified three-stage framework that produces a quantized DNN with balanced fault and attack robustness. Experiments on ResNet18, VGG16, EfficientNet, and Swin-Tiny in CIFAR-10, CIFAR-100, and GTSRB show consistent gains of up to 10.35\% in attack resilience and 12.47\% in fault resilience, without sacrificing clean and fault-free accuracy. Results also highlight a key asymmetric behavior that adversarial tuning slightly reduces fault resilience, while fault-aware fine-tuning restores and often enhances both resilience dimensions.

\section{Acknowledgments}
\scriptsize
This work was supported in part by the Estonian Research Council grant PUT PRG1467 "CRASHLESS“, EU Grant Project 101160182 “TAICHIP“, by the Deutsche Forschungsgemeinschaft (DFG, German Research Foundation) – Project-ID "458578717", and by the Federal Ministry of Research, Technology and Space of Germany (BMFTR) for supporting Edge-Cloud AI for DIstributed Sensing and COmputing (AI-DISCO) project (Project-ID "16ME1127").

\bibliographystyle{IEEEtran}
\bibliography{ref}

\end{document}